\documentclass[10pt,journal,compsoc]{IEEEtran}
\ifCLASSOPTIONcompsoc
  \usepackage[nocompress]{cite}
\else
  \usepackage{cite}
\fi
\ifCLASSINFOpdf
\else
\fi
\usepackage{times}
\usepackage{soul}
\usepackage{url}
\usepackage[utf8]{inputenc}
\usepackage{graphicx}
\usepackage{amsmath}
\usepackage{amsthm}
\usepackage{booktabs}
\usepackage{algorithm}
\usepackage{algorithmic}
\usepackage{lipsum}
\usepackage{amsfonts}
\usepackage{amsmath}
\begin{document}
%
\title{Towards Representation Identical Privacy-Preserving Graph Neural Network via Split Learning}
%
%
%
%

\author{Chuanqiang~Shan,
        Huiyun~Jiao,
        and~Jie~Fu
}

%
%

\markboth{Journal of \LaTeX\ Class Files,~Vol.~14, No.~8, August~2015}%
{Shell \MakeLowercase{\textit{et al.}}: Bare Demo of IEEEtran.cls for Computer Society Journals}
%



\IEEEtitleabstractindextext{%
\begin{abstract}
In recent years, the fast rise in number of studies on graph neural network (GNN) has put it from the theories research to reality application stage. Despite the encouraging performance achieved by GNN, less attention has been paid to the privacy-preserving training and inference over distributed graph data in the related literature. Due to the particularity of graph structure,  it is challenging to extend the existing private learning framework to GNN. Motivated by the idea of split learning, we propose a \textbf{S}erver \textbf{A}ided \textbf{P}rivacy-preserving \textbf{GNN} (SAPGNN) for the node level task on horizontally partitioned cross-silo scenario. It offers a natural extension of centralized GNN to isolated graph with max/min pooling aggregation, while guaranteeing that all the private data involved in computation still stays at local data holders. To further enhancing the data privacy, a secure pooling aggregation mechanism is proposed. Theoretical and experimental results show that the proposed model achieves the same accuracy as the one learned over the combined data. 
\end{abstract}

}

\maketitle

\IEEEdisplaynontitleabstractindextext

%
\IEEEpeerreviewmaketitle

\IEEEraisesectionheading{\section{Introduction}\label{sec:introduction}}

%
%
%
%
\IEEEPARstart{S}{ince} graph neural network (GNN) enables directly model the structure information of network topology, it has attracted significant interest recently, both from research and application perspectives \cite{wu2020comprehensive,zhou2020graph}. However, primarily due to business competition and regulatory restrictions, a wealth of sensitive graph-structured data that is held by different clients are unwilling to be shared, thus plaguing many practical applications, such as fraud detection over banks \cite{kurshan2020graph} and social network recommendation over platforms \cite{wu2021fedgnn}. 

Despite kinds of privacy preserving machine learning models have been successfully applied in data types like image \cite{hsu2020federated}, text \cite{ge2020fedner} and table \cite{wu2021fedgnn}, few works have concentrated on the domain of graph machine learning. For decentralized graph structure data, both nodes and edges are isolated, rendering most of the privacy learning methods designed for conventional datasets infeasible. 

\begin{figure}
	\centering
	\includegraphics[width=0.95\linewidth]{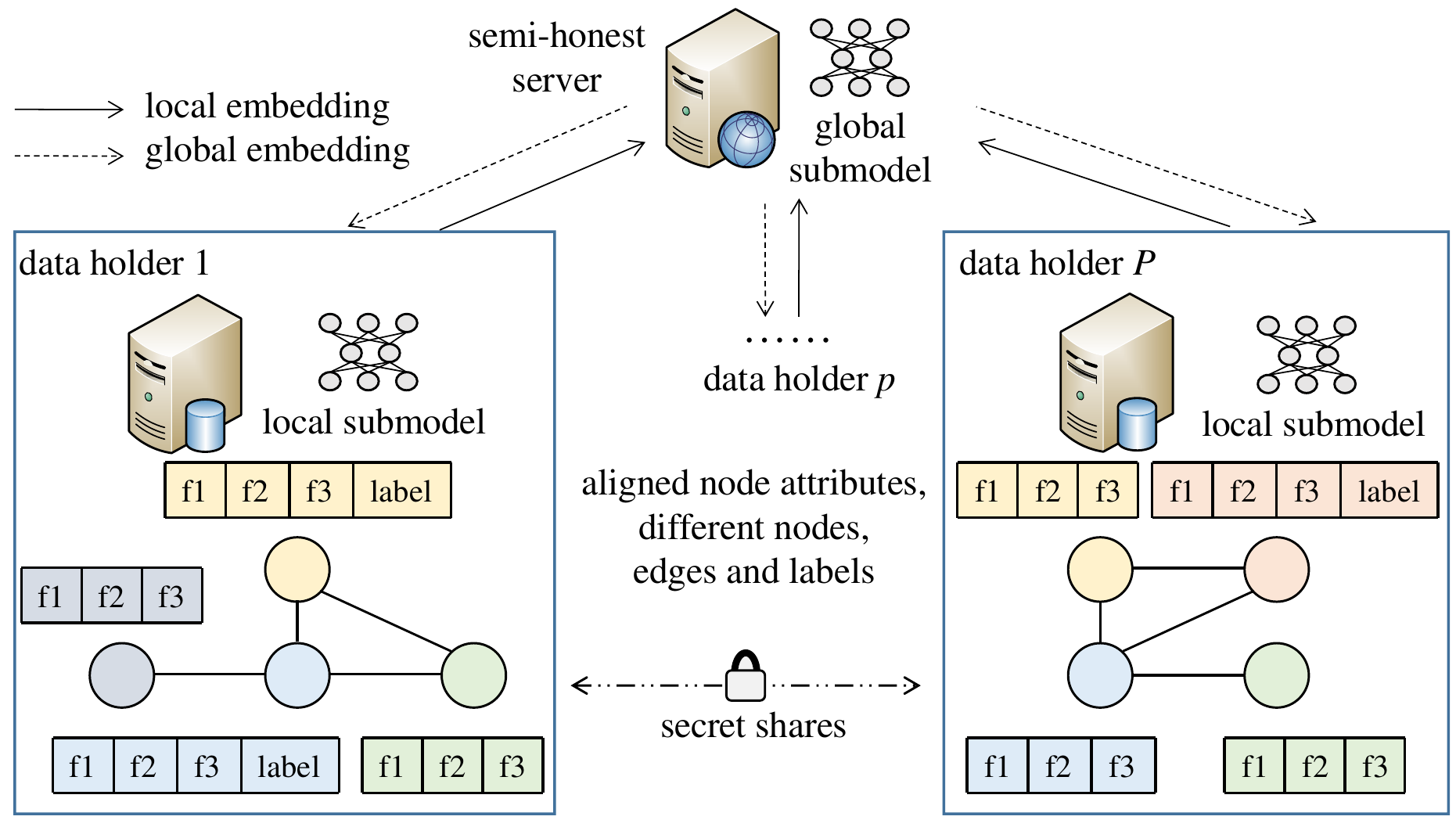}
	\caption{The proposed SAPGNN on horizontally partitioned data. The isolated data holders have the same feature domain (e.g., \{f1,f2,f3\}) and edge type, but differ in nodes, edges and labels.}
	\label{fig:fig1}
\end{figure}

In this work, we restrict attention to the problem of designing a privacy-preserving GNN for node classification task that allows performance intact in the setting of \textit{horizontally partitioned graph dataset}, which means the attributes of nodes and edges are aligned. As illustrated in Fig.\ref{fig:fig1}, we consider the scenario where several data holders which store private subgraphs access to one \textit{semi-honest} (a.k.a. honest-but-curious) server. Each local subgraph contains sensitive information about nodes, edges, attributes and labels. The semi-honest server assumption means the server will follow protocol honestly, but it attempts to infer as much information as possible from received messages. 
In view of the fact that one node may interact with the same entity at several platforms, unlike previous work, we consider a more general scenario where overlapped nodes and edges exist among  subgraphs. 

To address the decentralized graph learning issue under privacy constraint, motivated by the ideas of \textit{split learning} \cite{vepakomma2018split} and \textit{horizontal federated learning} \cite{aono2017privacy}, we propose a \textbf{S}erver \textbf{A}ided \textbf{P}rivacy-preserving \textbf{GNN} (SAPGNN), where each GNN layer is divided into two sub-models: the local model includes all the private data related computation to generate local node embedding, whereas the global model calculates global embedding by aggregating all local embedding. By this, the isolated neighborhood can be collaboratively utilized, and the receptive field can be enlarged by stacking multiple layers. Most importantly, when employing a pooling aggregator with proper update function, SAPGNN can generate identical node representation compared to the one learned over the combined graph. 

The main contributions of this paper are summarized as follows:
\begin{itemize}
	\item[$\bullet$] We present a novel SAPGNN framework for training privacy-preserving GNN in horizontally partitioned data setup. To the best of our knowledge, it is the first GNN learning paradigm that is capable of generating the same node embedding as the centralized counterpart.
	\item[$\bullet$] We analyse the privacy and overhead of the proposed SAPGNN. A secure pooling mechanism instead of a naive global pooling aggregator is proposed to further protect privacy from the semi-honest adversaries of server.
	\item[$\bullet$] Experimental results on three datasets demonstrate the accuracy and macro-f1 of SAPGNN surpass the one learned over isolated data, and also comparable to the state-of-the-art approach, especially in the setting of I.I.D. label distribution.
\end{itemize}

This paper is organized as follows: Section 2 and 3 introduce recent works on privacy preserving GNN learning paradigms, notations as well as preliminaries; In
Section 4 we describe and discuss our proposed SAPGNN framework in detail; These are followed by the experiments in Section 5; and finally Section 6
provides conclusion discussions and outlook.

\section{Related Works}	
To tackle the privacy-preserving node classification problem over decentralized graph data, some methods have recently been investigated to train a global GNN collaboratively on various split types of dataset. 

First, two learning paradigms named PPGNN \cite{zhou2020privacy} and ASFGNN \cite{zheng2021asfgnn} were proposed based on split learning for vertically and horizontally split datasets respectively.
Both of them alleviate isolation by firstly training local GNN models over private graphs and then learning global embedding at an assistant semi-trusted third party. As the graph topology is still exploited locally, the model performance may be substantially reduced when the dataset is largely decentralized. 
More recently, LPGNN \cite{sina2020practical} was developed to reduce communication overhead under the assumption that the  server has accessed global graph topology except private node attributes.  Despite its potential, this precondition is not always satisfactory since releasing topology to server may lead to privacy disclosure risk.  We show the comparison of these methods in Table \ref{tab:split}.

\begin{table}
	\caption{The comparison of data partition manners}
	\centering
	\label{tab:split}
	\begin{tabular}{lccc}
		\toprule
		Model  & Nodes & Edges & Features \\
		\midrule
		PPGNN       & aligned  & not limited & not limited\\
		ASFGNN       & different  & different & aligned \\
		LPGNN       & not limited  & shared to server& aligned \\
		SAPGNN       & not limited & not limited& aligned\\
		\bottomrule
	\end{tabular}
\end{table}

From the perspective of application, \cite{wu2021fedgnn} proposes a GNN-based privacy-preserving recommendation framework for the decentralized learning from user-item graph. \cite{hefedgraphnn} presents an open-source federated learning system and gives important insights into the federated GNN training over non-I.I.D. molecular datasets.

The nice property of our proposed SAPGNN is that it generates the same
node embedding as the centralized GNN without having access to
the raw data stored at other data holders. Unlike previous works, it
can achieve the same accuracy as the one learned over the combined
data for isolated datasets. In addition, it relaxes the constraints on the partition manners of both nodes and edges.

\section{Preliminaries}
For clarity, we summarize all the notations used in this paper in Table \ref{tab:Notations}.
\begin{table}
		\caption{ Notations and descriptions.}
	\centering
	\begin{tabular}{cl}
		\toprule
		Not.  & Descriptions \\
		\midrule
		$G^p$       & local graph of data holder $p$  \\
		$V^p$       & nodes of   data holder $p$  \\
		$E^p$       & edges of  data holder $p$\\
		$P$       & total number of data holders  \\
		$\mathcal{P}$       & set of data holders \\
		$L$       & total number of layers\\
		$L_p$       & local loss at data holder $p$\\
		$\mathcal{L}$       & total loss of all data holders\\
		$\mathcal{N}^p(v)$       & neighbour of node $v$ at data holder $p$\\
		$\mathbf{h}_{v}^{(l)}$      & input embedding of
		node $v$ at the $1$-th layer\\
		&and global embedding at the $(1-1)$-th layer\\
		$\mathbf{m}_{v u}^{(l,p)}$      & message of the edge connected to node $u$ and $v$\\
		& at data holder $p$\\
		$\mathbf{m}_{v}^{(l,p)}$      & local aggregation of node $v$ at data holder $p$\\
		$\mathbf{t}_{v}^{(l,p)}$&local embedding of node $v$ at data holder $p$\\
		$\mathbf{m}_{v}^{(l)}$&global aggregation of node $v$ at server\\
		$\boldsymbol{\rho}^{(l)}(\cdot)$& message construction function at layer $l$\\
		$\boldsymbol{\phi}^{(l)}(\cdot)$& local vertex update function at layer $l$\\
		$\boldsymbol{\varphi}^{(l)}(\cdot)$& global vertex update function at layer $l$\\
		$\bigoplus$&XOR operator\\
		$\mathbb{Z}_{b}$&Nonnegative integer set not greater than $b$\\
		$\langle\cdot\rangle$&encryption using additive sharing\\
		$\langle\cdot\rangle_{\oplus}$&encryption using  boolean sharing\\
		$\mathbf{\Theta}$&model weights\\
		$\mathcal{G}_\mathbf{\Theta}^{(l,p)}$&the gradient of local weights at data holder $p$\\
		$W$&the data size of local model weights\\
		$b$& the length of node embedding\\
		$d$&s the data size of the value of each weight\\
		$N$&the number of nodes from all local graphs\\
		$\mathbf{W}$&weights of linear transformer matrix\\
		$q$& the label distribution ratio\\
		\bottomrule
	\end{tabular}
	\label{tab:Notations}
\end{table}
\subsection{Graph representation learning}
Let $G=(V,E)$ defines a graph with vertex set $V$ and edge set $E$. Most existing $L$-layer stacked GNN models can be viewed as a special case of message passing architecture \cite{gilmer2017neural}. Specifically, at the $l$-th layer, the message passing on node $v \in V$ and its neighborhood set $\mathcal{N}(v)$ can be composed of three steps:
\begin{align}\label{mf}
	\mathbf{m}_{v u}^{(l)} & = \boldsymbol{\rho}^{(l)}(\mathbf{h}_{v}^{(l)}, \mathbf{h}_{u}^{(l)}, \mathbf{h}_{e_{v u}}^{(l)}), u \in \mathcal{N}(v),
\end{align}
\begin{align}\label{rf}
	\mathbf{m}_{v}^{(l)} & = \boldsymbol{\zeta}^{(l)}(\{\mathbf{m}_{v u}^{(l)} \mid u \in \mathcal{N}(v)\}),
\end{align}
\begin{align}\label{uf}
	\mathbf{h}_{v}^{(l+1)} & = \boldsymbol{\phi}^{(l)}(\mathbf{h}_{v}^{(l)}, \mathbf{m}_{v}^{(l)}),
\end{align}
where $\boldsymbol{\rho}^{(l)}(\cdot)$ in (\ref{mf}) is a {\it message construction function} defined on each edge connected to $v$. The message $\mathbf{m}_{v u}^{(l)}$ is constructed by combining the edge feature $\mathbf{{h}}_{e_{v u}}^{(l)}$ with the features of its incident nodes $\mathbf{h}_{v}^{(l)}$ and $\mathbf{h}_{u}^{(l)}$; The {\it message aggregation function} $\boldsymbol{\zeta}^{(l)}(\cdot)$ in (\ref{rf}) calculates $\mathbf{m}_{v}^{(l)}$ by aggregating the feature of incoming finite unordered message set. The function is usually designed as a permutation invariant set function to guarantee the invariance/equivariance to isomorphic graph, popular choices include mean \cite{hamilton2017inductive}, pooling \cite{li2019deepgcns}, sum \cite{xu2018powerful} and attention \cite{velickovic2018graph}. {\it Vertex update function} $\boldsymbol{\phi}^{(l)}(\cdot)$  in (\ref{uf}) updates the node feature according to its own feature $\mathbf{h}_{v}^{(l)}$ and the aggregated message $\mathbf{m}_{v}^{(l)}$. 
Lastly, the node representations are applied to loss functions for specific downstream tasks, e.g., node or graph classification \cite{kurshan2020graph,xu2018powerful}, link prediction \cite{wu2021fedgnn}, etc.

\subsection{Split learning}
Unlike federate learning \cite{yang2019federated} where each client trains an entire replica of model, the keynote of split learning is splitting the execution of a model on a per-layer basis between clients and aided server \cite{vepakomma2018split,gupta2018distributed}. In principle, each data holder first finishes the private data related calculation up to a {\it cut layer}, then the outputs are sent to another entity for subsequent computation. After the forward propagation, the gradients are computed based on loss function and backward propagated. Throughout the training or inference process, data privacy is guaranteed by the fact that raw data only participates in local computation and will never be accessed by others. Both theoretical analysis \cite{singh2019detailed} and practical application \cite{gao2020end} compare the efficiency and effectiveness of federated learning and split learning, and show the potential of both methods to design private decentralized learning procedures.
For more details and advances, we refer to the reference \cite{kairouz2019advances} and the website\footnote{\url{ https://splitlearning.github.io/}}.

\subsection{Secret sharing}
Our proposed model employs n-out-of-n secret sharing schemes to recover privacy from secret shares \cite{shamir1979share,demmler2015aby}. In particular, when client $p$ wants to share a $b$-bit value $x$ to $\mathcal{P}$ parties,  it firstly generates and sends $\{x_i\in  \mathbb{Z}_{2^b} \mid i\in\mathcal{P}, i\neq p\}$ uniformly at random to each client $i$ and generates $x_p$ that satisfies $x=\sum_{i\in\mathcal{P}}x_i$ mod $2^b$ for additive sharing and  $x=\bigoplus_{i\in\mathcal{P}}x_i$ for boolean sharing, respectively. Accordingly, $x$ can be reconstructed at any entity by gathering all shared values. Secret sharing has become one popular basis of advanced secure multi-party computation frameworks \cite{patra2020aby2,byali2020flash} and been applied to many privacy preserving machine learning algorithms, such as secure aggregation \cite{bonawitz2017practical}, embedding generation \cite{zhou2020privacy}, and secure computation \cite{mohassel2020practical}. For clarity, we denote additive sharing by $\langle\cdot\rangle$ and boolean sharing by $\langle\cdot\rangle_{\oplus}$ in the following.


\section{The Proposed SAPGNN Framework}
In this section, we describe the proposed SAPGNN framework that has the ability to keep accuracy intact compared to the counterpart learned over the combined graph. 
The learning paradigm consists of \textit{parameter initialization}, \textit{forward propagation}, \textit{back propagation} and \textit{local parameter fusion}. At last, we give a discussion about additional overhead and data privacy in the presence of semi-honest adversaries.

\subsection{Parameter initialization}
First of all, the participated data holders and server build  pair-wise secure channels for all sequential communication to ensure data integrity.
Recall that all the nodes from local graphs share the same feature domain. Inspired by horizontal federated learning \cite{aono2017privacy}, local models at all data holders are initialized  by the same weights to keep identical model behavior. This can be easily implemented by sharing the same initialization approach and random seed. Additionally, the shared parameters also include: (1) training hyperparameters that are shared among data holders and server, (2) hashed node index list that only shared to server. The hashed index list is used to index and distinguish nodes from all local graphs to hide the raw index information from the server.
As for the server, it requires randomly initialization of global model weights to generate global embedding.

\subsection{Forward Propagation}
\begin{figure*}
	\centering
	\includegraphics[width=0.85\linewidth]{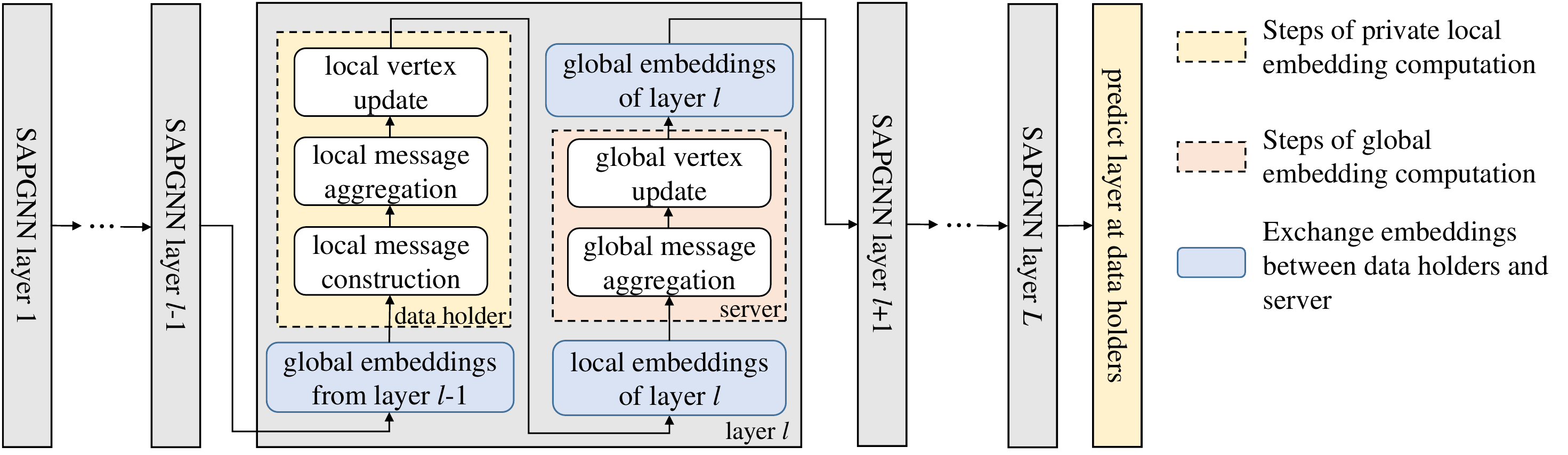}
	\caption{Forward propagation of SAPGNN. At each layer, local embeddings are first computed by message passing architecture over local graph at each data holder side. Then the server obtains global embeddings via global message aggregation and vertex update steps at the server side. At last, label prediction is conducted individually at each data holder.}
	\label{fig:fig2}
\end{figure*}
As illustrated in Fig.\ref{fig:fig2}, in order to protect data privacy (i.e., node attributes, edge information and node labels) while exploiting all isolated graph information, we design a \textit{modified message passing architecture} in the manner of layer-wise split learning. To be specific, the forward steps at each layer can be divided into two steps: it first calculates local embedding at each data holder individually with private data. Then, the semi-honest server collects non-private local embeddings to compute global embedding. In the end, the output of the last layer is sent to label prediction and loss computation functions. 
\subsubsection{Private local embedding computation} In line of the message passing architecture, each data holder first \textit{constructs local message} as
\begin{align}\label{mfl}
	\mathbf{m}_{v u}^{(l,p)} & = \boldsymbol{\rho}^{(l)}(\mathbf{h}_{v}^{(l)}, \mathbf{h}_{u}^{(l)}, \mathbf{h}_{e_{v u}}^{(l)};\mathbf{\Theta}_\rho^{(l)}), u \in \mathcal{N}^{p}(v),
\end{align}
where $\mathcal{N}^{p}(v)$ denotes the neighbor node set at the local graph of data holder $p$, $\mathbf{\Theta}_\rho^{(l)}$ is the parameters of function $\rho^{(l)}$.

The next step is  {\it local message aggregation}.
Suppose that aggregation is conducted over a combined graph from all data holders.
Since the same edge may simultaneously appear at several data holders, it will lead to count the same node multiple times when sum \cite{xu2018powerful}, mean \cite{hamilton2017inductive} and degree-based \cite{kipf2017semi} aggregators are employed. 
Fortunately, max/min pooling aggregator tackles this problem naturally, therefore we will complete the decentralized learning paradigm based on the pooling aggregator. 
Taking max pooling as an example, each data holder $p$ aggregates messages over local neighbors by
\begin{equation}
	\begin{aligned}	\label{rfl}
		\mathbf{m}_{v}^{(l,p)} = \mathrm{max}(\{\mathbf{m}_{v u}^{(l,p)} \mid u \in \mathcal{N}^{p}(v)\}).	
	\end{aligned}
\end{equation}

After local aggregation, each data holder calculates the local node embeddings based on the node feature $\mathbf{h}_{v}^{(l)}$ and aggregated neighbor feature $\mathbf{m}_{v}^{(l,p)}$ via {\it local vertex update} function
\begin{align}\label{ufl}
	\mathbf{t}_{v}^{(l,p)} = \begin{cases}
		\boldsymbol{\phi}^{(l)}(\mathbf{h}_{v}^{(l)}, \mathbf{m}_{v}^{(l,p)};\mathbf{\Theta}_\phi^{(l)})& \text{ if } v \in V^p \\
		-\mathbf{inf}& \text{ if } v \notin V^p 
	\end{cases},
\end{align}
where $\mathbf{inf}$ denotes the vector whose elements are all infinitesimals. The local embeddings $\mathbf{t}_{v}^{(l,p)}$ hide raw information of local graph, hence it can be sent to server for further global computation.

\subsubsection{Global embedding computation} This step consists of {\it global aggregation} and {\it vertex update}. Concretely, the server first aggregates local node embeddings from all data holders with the same pooling function to \eqref{rfl} by
\begin{equation}
	\begin{aligned}
		\label{rfg}
		\mathbf{m}_{v}^{(l)}  = \mathrm{max}(\{\mathbf{t}_{v }^{(l,p)} \mid p \in \mathcal{P}\}).
	\end{aligned}
\end{equation}
After that, the server transforms the aggregated embedding $\mathbf{m}_{v}^{(l)}$ to compute global node representation of layer $l$ as
\begin{equation}
	\begin{aligned}\label{ufg}
		\mathbf{h}_{v}^{(l+1)}  = \boldsymbol{\varphi}^{(l)}( \mathbf{m}_{v}^{(l)};\mathbf{\Theta}_\varphi^{(l)}),
	\end{aligned}
\end{equation}
To meet the various design space for GNN \cite{you2020design},  the combination of linear transformer, batchnorm, activation and dropout can be incorporated into the vertex update function $\boldsymbol{\varphi}^{(l)}(\cdot)$ to enhance model capacity.

Note that the result of pooling aggregation in \eqref{rfl} and \eqref{rfg} only depends on the element-wise maximum. In order to follow the same behavior of centralized GNN layer, the global aggregated result $\mathbf{m}_{v}^{(l)}$ (i.e., the left side of \eqref{satisfy}) should be identical to
that aggregated at all neighbors of combined graph (i.e., the right side of \eqref{satisfy}), which can be formulated as
\begin{equation}
	\begin{aligned}\label{satisfy}
		\mathrm{max}(\{\boldsymbol{\phi}^{(l)}(\mathbf{h}_{v}^{(l)}, \mathrm{max}(\{\mathbf{m}_{v u}^{(l,p)} \!\mid\! u \!\in\! \mathcal{N}^{p}(v)\})) \!\mid\! p \!\in\! \mathcal{P}\})\\
		=\boldsymbol{\phi}^{(l)}(\mathbf{h}_{v}^{(l)}, \mathrm{max}(\{\mathbf{m}_{v u}^{(l,p)} \!\mid\! u \!\in\! \mathcal{N}^{p}(v),p \!\in\! \mathcal{P}\}))
	\end{aligned}
\end{equation}
To satisfy the equation above, the constraints of \textit{local updates function} $\boldsymbol{\phi}^{(l)}(\cdot)$ can be given in the following:

\textbf{Proposition}[Constraints of local updates function]
	When the aggregation function is element-wise max, each element of the output of local update function  $\boldsymbol{\phi}(\mathbf{h}_{v}^{(l)}, \mathbf{m}_{v}^{(l,p)})$ should monotonically increase with each increased element of $\mathbf{m}_{v}^{(l,p)}$, e.g. $\boldsymbol{\phi}(\cdot)$ can be chosen from $\mathbf{h}_{v}^{(l)}\|\mathbf{m}_{v}^{(l,p)}$, $\mathbf{h}_{v}^{(l)}+ \mathbf{m}_{v}^{(l,p)}$ and $(\mathrm{ReLU}(\mathrm{MLP}(\mathbf{h}_{v}^{(l)})))*\mathbf{m}_{v}^{(l,p)}$, where $\|$ denotes concatenation, $*$ denotes element-wise multiplication, $\mathrm{MLP}$ denotes multilayer perceptron.
\begin{proof}
	Denote the results of element-wise max aggregation at the local neighbor information as 
	\begin{equation}
		\begin{aligned}\label{pf1}
			\mathbf{m}_{v}^{*(l,p)} & = \mathrm{max}\left(\left\{\mathbf{m}_{v u}^{(l,p)} \mid u \in \mathcal{N}^{p}(v)\right\}\right),
		\end{aligned} 
	\end{equation}
	and the entire neighbor information as 
	\begin{equation}
		\begin{aligned}\label{pf2}
			\mathbf{m}_{v}^{*(l)} & = \mathrm{max}\left(\left\{\mathbf{m}_{v u}^{(l,p)} \mid u \in \mathcal{N}^{p}(v),p \in \mathcal{P}\right\}\right)\\
			& = \mathrm{max}\left(\left\{	\mathbf{m}_{v}^{*(l,p)} \mid p \in \mathcal{P}\right\}\right),
		\end{aligned}
	\end{equation}
	respectively. Incorporating \eqref{pf1} and \eqref{pf2}, equation \eqref{satisfy} in the paper can be simplified as 
	\begin{equation}\begin{aligned}\label{pf3}
			&\mathrm{max}(\{\boldsymbol{\phi}^{(l)}(\mathbf{h}_{v}^{(l)}, \mathbf{m}_{v}^{*(l,p)}) \mid p \in \mathcal{P}\})\\
			=&\boldsymbol{\phi}^{(l)}(\mathbf{h}_{v}^{(l)}, \mathbf{m}_{v}^{*(l)})\\
			=&\boldsymbol{\phi}^{(l)}(\mathbf{h}_{v}^{(l)}, \mathrm{max}(\{	\mathbf{m}_{v}^{*(l,p)} \mid p \in \mathcal{P}\}))
	\end{aligned}\end{equation}
	Omits the layer index $l$ and node index $v$, denote $\boldsymbol{\phi}'(\mathbf{m})=\boldsymbol{\phi}(\mathbf{h}, \mathbf{m})$, the above equation reduces to
	\begin{equation}\begin{aligned}\label{pf4}
			\mathrm{max}(\{\boldsymbol{\phi}'(\mathbf{m}_{v}^{*(p)})\! \mid \!p \in \mathcal{P}\})=\boldsymbol{\phi}'(\mathrm{max}(\{	\mathbf{m}_{v}^{*(p)} \!\mid\! p \in \mathcal{P}\}))	
	\end{aligned}\end{equation}		
	Obviously, according to the property of max function, equation \eqref{pf3} holds if and only if $\partial{\boldsymbol{\phi}^{(l)}(\mathbf{h},\mathbf{m})}/\partial{m}\geq0$ for $\forall \mathbf{h} \in \mathbb{R}^{d_1}$ and each element $m$ of vector $\mathbf{m}  \in \mathbb{R}^{d_2}$, i.e., each element of the output of  $\boldsymbol{\phi}^{(l)}(\mathbf{h},\mathbf{m})$ should monotonically increase with the increasing of each element of $\mathbf{m}$. 
\end{proof}
	
When the global embeddings of all nodes at layer $l$ have been obtained by \eqref{ufg}, the server distributes them to each data holder according to the node list of local graph for the forward propagation of the next layer. This process is conducted iteratively until the last layer $L$.
\subsubsection{Private local loss computation}  
When the global node embeddings of the last layer $L$ have been computed, each data holder predicts labels based on the  embeddings by
\begin{align}\label{pl}
	\widetilde{y}_v^p=\boldsymbol{\psi}(\mathbf{h}_{v}^{(L+1)};\mathbf{\Theta}_\psi),
\end{align}
the local loss at data holder $p$  over local training node labels can be then computed by
\begin{align}\label{lossp}
	L_p=-\sum_{i=1}^{N_p}\boldsymbol{\Xi}(y_v^p,\widetilde{y}_v^p)
\end{align}
respectively, where $N_p$ is the number of labels at data holder $p$, $\boldsymbol{\Xi}$ is the loss function, such as cross-entropy for a
classification task and mean square loss for a regression task. 

To summarize, the forward propagation algorithm is given in Algorithm \ref{alg:ag1}. When the forward propagation is finished, model weights can be updated by the back propagation procedure outlined in what follows.

\begin{algorithm}[tb]
	\caption{Forward propagation of SAPGNN learning algorithm}
	\label{alg:ag1}
	\textbf{Input}: local graph set $\{G^p(V^p, E^p), p \in \mathcal{P}\}$ on data holder $p$ and node features $\mathbf{h}_v$ for $v \in V^p$; all node set $V=\bigcup_p \{V^p\}$; labeled node set $\bar{V}=\bigcup_p \{\bar{V}^p\}$; the number of layers $L$;\\ 
	\textbf{Output}: Label predictions $\{y_v^p\mid \forall v \in \bar{V}, \forall p \in \mathcal{P}\}$ and loss $\{L_p\mid \forall p \in \mathcal{P}\}$ on each data holder $p$;
	\begin{algorithmic}[1] 
		\FOR{$l=1$ to $L$}
		\FOR{$p \in \mathcal{P}$  in parallel} 
		\STATE \textbf{Data holder:} calculates local node embedding $\{\mathbf{t}_{v}^{(l,p)} \mid \forall v \in V\}$ by (\ref{mfl})-(\ref{ufl}) and sends to server.
		\ENDFOR
		\STATE \textbf{Server}: combines the local embeddings to calculate global  embedding $\{\mathbf{h}_{v}^{(l+1)} \mid \forall v \in V\}$ by (\ref{rfg}) and (\ref{ufg}), then distributes them back based on node lists $V^p$. 
		\ENDFOR
		
		\STATE  \textbf{Data holder:} private label prediction and loss computation $\{y_v^p, L_p\mid \forall v \in \bar{V}, \forall p \in \mathcal{P}\}$ by (\ref{pl}) and (\ref{lossp}).
	\end{algorithmic}
\end{algorithm}

\subsection{Back Propagation}
Recall that the local part (i.e., the local embedding and loss related computation at data holders) and the global part (i.e., the global embedding related computation at server side) at each layer are spatially isolated. According to the chain rule of derivation, the entire model can be updated iteratively through communicating intermediate gradients between data holder and server. Herein, the gradient of local model weights are computed individually and then secretly aggregated for update. In the following, we give the computation and communication of back propagation procedure in detail. 
\subsubsection{Individual back propagation of predict layer}
As the bridge of the final node embedding and the model output, the weight of predict function $\mathbf{\Theta}_\psi$ at data holder $p$ can be first learned by gradient descent through minimizing local loss $\partial{L_p}/\partial{\mathbf{\Theta}_\psi}$ individually. 
After that, the data holder computes the gradient of loss with respect to the input of predict function $\partial{L_p}/\partial{\mathbf{h}_{v}^{(L+1)}}$ and then sends it to the server for subsequent back propagation.

\subsubsection{Back propagation of each SAPGNN layer}
Due to the property of derivation, the gradient of entire loss $\mathcal{L}$ with respect to the output of the last layer $\mathbf{h}_{v}^{(L+1)}$ can be computed as
\begin{align}\label{derivative}
	\frac{\partial{\mathcal{L}}}{\partial{\mathbf{h}_{v}^{(L+1)}}}=\frac{\partial{\sum_{p}L_p}}{\partial{\mathbf{h}_{v}^{(L+1)}}}=\sum_{p}\frac{\partial{L_p}}{\partial{\mathbf{h}_{v}^{(L+1)}}}, 
\end{align}
while for the $l$-th layer ($l<L$), based on the derivation of max function, the gradient of loss $\mathcal{L}$ with respect to the input embedding $\mathbf{h}_{v}^{(l)}$ can be decomposed as 
\begin{align}\label{derivative0}
	\frac{\partial{\mathcal{L}}}{\partial{\mathbf{h}_{v}^{(l)}}}=\underset{= \mathrm{P1} }{\underbrace{\frac{\partial{\mathcal{L}}}{\partial{\mathbf{h}_{v}^{(l+1)}}}}}\underset{= \mathrm{P2} }{\underbrace{\frac{\partial{\mathbf{h}_{v}^{(l+1)}}}{\partial{\mathbf{m}_{v}^{(l)}}}}}\sum_{p} \underset{= \mathrm{P3} }{\underbrace{\frac{\partial{\mathbf{m}_{v}^{(l)}}}{\partial{\mathbf{t}_{v}^{(l,p)}}}}}\underset{= \mathrm{P4} }{\underbrace{\frac{\partial{\mathbf{t}_{v}^{(l,p)}}}{\partial{\mathbf{h}_{v}^{(l)}}}}}, 
\end{align}
where 
part $\mathrm{P1}$ denotes the gradient of loss $\mathcal{L}$ with respect to the output global embedding of layer $l$, $\mathrm{P2}$ denotes the gradient of global embedding  $\mathbf{h}_{v}^{(l+1)}$ with respect to the result of global aggregation $\mathbf{m}_{v}^{(l)}$, $\mathrm{P3}$ denotes the gradient of $\mathbf{m}_{v}^{(l)}$ with respect to the input of global model $\mathbf{t}_{v}^{(l,p)}$. Obviously, both $\mathrm{P2}$ and $\mathrm{P3}$ can be computed at the server side. Part $\mathrm{P4}$ denotes  the gradient of data holder output $\mathbf{t}_{v}^{(l,p)}$ with respect to the input embedding  $\mathbf{h}_{v}^{(l)}$, this can be obtained at each data holder individually. Therefore, according to (\ref{derivative0}), the gradients can be back propagated layer by layer recursively. At each layer, the propagation is first carried out globally at the server side and then locally and parallelly at each data holder side. 

\subsubsection{Global back propagation at server side.}

The server first obtains $\partial{\mathcal{L}}/\partial{\mathbf{h}_{v}^{(l+1)}}$ by summing received gradients $\{\partial{L_p}/\partial{\mathbf{h}_{v}^{(l+1)}} \mid p \in \mathcal{P}\}$ from all data holders, and
then computes the derivation with respect to global model weights $\mathbf{\Theta}_\varphi^{(l)}$ and local embedding $\mathbf{t}_{v }^{(l,p)}$ for every $p \in \mathcal{P}$:
\begin{align}\label{derivative1}
	\frac{\partial{\mathcal{L}}}{\partial{\mathbf{\Theta}_\varphi^{(l)}}}=\frac{\partial{\mathcal{L}}}{\partial{\mathbf{h}_{v}^{(l+1)}}}\frac{\partial{\mathbf{h}_{v}^{(l+1)}}}{\partial{\mathbf{\Theta}_\varphi^{(l)}}}
\end{align}
\begin{align}\label{derivative2}
	\frac{\partial{\mathcal{L}}}{\partial{\mathbf{t}_{v }^{(l,p)}}}=\frac{\partial{\mathcal{L}}}{\partial{\mathbf{h}_{v}^{(l+1)}}}\frac{\partial{\mathbf{h}_{v}^{(l+1)}}}{\partial{\mathbf{m}_{v }^{(l)}}}\frac{\partial{\mathbf{m}_{v}^{(l)}}}{\partial{\mathbf{t}_{v }^{(l,p)}}}
\end{align}
respectively. The result of (\ref{derivative2}) is sent to corresponding data holder $p$ for the sequential local back propagation.

\subsubsection{Local back propagation at data holder side.} 
The gradient of loss $\mathcal{L}$ with respect to local weights set $\mathbf{\Theta}^{(l)}\in\{\mathbf{\Theta}_\rho^{(l)},\mathbf{\Theta}_\phi^{(l)}\}$ at data holder $p$ can be expressed as
\begin{align}\label{derivative3}
	\mathcal{G}_\mathbf{\Theta}^{(l,p)}= \left.\frac{\partial{\mathcal{L}}}{\partial{\mathbf{\Theta}^{(l)}}}\right|_{p}=\frac{\partial{\mathcal{L}}}{\partial{\mathbf{t}_{v}^{(l,p)}}}\frac{\partial{\mathbf{t}_{v}^{(l,p)}}}{\partial{\mathbf{\Theta}^{(l)}}}.
\end{align}
According to $\mathrm{P4}$ of (\ref{derivative0}), each data holder also needs to calculate and send the gradient of  output local embedding $\mathbf{t}_{v}^{(l,p)}$ with respect to input node embedding $\mathbf{h}_{v}^{(l)}$ to server.



\begin{algorithm}[tb]
	\caption{Back propagation and weights update of SAPGNN framework}
	\label{alg:ag2}
	\begin{algorithmic}[1] 
		\STATE $\Box$ \textbf{Step 1:} Back propagation of predict layer
		\FOR{$p \in \mathcal{P}$  in parallel} 
		\STATE \textbf{Data holder $p$:} computes $\partial{L_p}/\partial{\mathbf{\Theta}_\psi^{(p)}}$, computes and sends $\partial{L_p}/\partial{\mathbf{h}_{v}^{(L+1)}}$ to server.
		\ENDFOR
		\STATE $\Box$ \textbf{Step 2:} Back propagation of SAPGNN layer $l$
		\FOR{$l=L$ to $1$}
		\STATE \textbf{Server:} computes $\mathrm{P1}$ by (\ref{derivative}) if $l=L$ or (\ref{derivative0}) if  $l<L$, computes $\mathrm{P2}$ and $\mathrm{P3}$, computes gradient of global model weights by (\ref{derivative1}).
		\FOR{ $p \in \mathcal{P}$  in parallel} 
		\STATE \textbf{Data holder:} computes gradient of local model weights $\mathcal{G}^{(l,p)}$ by (\ref{derivative3}), computes and sends $\partial \mathbf{t}_{v}^{(l,p)}/ \partial \mathbf{h}_{v}^{(l)}$ to server. 
		\ENDFOR
		\ENDFOR
		\STATE $\Box$ \textbf{Step 3:} Weight update
		\FOR{\textbf{Data holder}  $p \in \mathcal{P}$  in parallel:} 
		\STATE locally generates $\{\langle\mathcal{G}_\mathbf{\Theta}^{(l,p)}\rangle_j\mid {j\in\mathcal{P}},{l\in{L}}\}$ and distributes  $\{\langle\mathcal{G}_\mathbf{\Theta}^{(l,p)}\rangle_j\mid{l\in{L}}\}$ to data holder $j$. 
		\STATE computes $\langle\mathcal{G}_\mathbf{\Theta}^{(l)}\rangle_p=\sum_{j}\langle\mathcal{G}_\mathbf{\Theta}^{(l,j)}\rangle_p$ and sends it to other data holders.
		\STATE reconstructs $\mathcal{G}_\mathbf{\Theta}^{(l)}=\sum_{p}\langle\mathcal{G}_\mathbf{\Theta}^{(l)}\rangle_p$ and updates local weights via gradient descent.
		\ENDFOR
		\STATE \textbf{Server:} updates global weights via gradient descent.
	\end{algorithmic}
\end{algorithm}

\subsection{Weights update}
As described above, the model weights of SAPGNN are spatially divided into two categories: global submodel weights $\{\mathbf{\Theta}_\varphi^{(l)} \mid l \in \mathcal{L}\}$ held by server and local submodel weights $\mathbf{\Theta}\in\{\mathbf{\Theta}_\rho^{(l)},\mathbf{\Theta}_\phi^{(l)},\mathbf{\Theta}_\psi \mid l \in \mathcal{L}\}$ held by data holders.
\subsubsection{Update of global model weight.} When the corresponding gradients have been obtained by (\ref{derivative1}), the global weights can be \textit{directly} updated through gradient descent.
\subsubsection{Update of local model weight.} To keep the isolated local weights of all data holders identical during training, the corresponding local gradients should be \textit{federally} aggregated at all data holders respectively, such as secure aggregation \cite{bonawitz2017practical} or homomorphic
encryption \cite{aono2017privacy}. Taking secure aggregation as an example, let the gradients of weights be aggregated at data holder $p$ as $\mathcal{G}_\mathbf{\Theta}^p$. Each data holder first secretly shares local gradient $\langle\mathcal{G}_\mathbf{\Theta}^p\rangle_i$ to the data holder $i$ and then sums up the shares by
\begin{align} \langle\mathcal{G}_\mathbf{\Theta}\rangle_p=\mathrm{sum}(\{\langle\mathcal{G}_\mathbf{\Theta}^j\rangle_p\mid j\in\mathcal{P}\}). 
\end{align}
After that, each data holder reconstructes the entire gradients for update by gathering aggregated results from others by \begin{align}\mathcal{G}_\mathbf{\Theta}=\mathrm{sum}(\{\langle\mathcal{G}_\mathbf{\Theta}\rangle_p\mid p\in\mathcal{P}\}). 
\end{align}
Note that during this procedure, each data holder only accesses the secret shares and  reconstructed entire gradients, whereas the server knows nothing about local gradients.

\subsection{Discussion of security and overhead}
\subsubsection{Data privacy}
In our proposed learning paradigm, data privacy can be guaranteed by the following reasons:
\begin{itemize}
	\item[$\bullet$] All aforementioned private data (including node attributes, edge information, labels and local model gradients) related computations are carried out by data holders locally. From the perspective of semi-honest server, only the hashed node lists of local graph, local embedding computed at (\ref{ufl}) and global model are observable. Therefore, our SAPGNN is secure against semi-honest adversaries.
	\item[$\bullet$] The only sensitive messages observed by data holders are the secret shares of gradients of local model weights. Since the gradients are split by n-out-of-n secret sharing algorithm, raw data can be reconstructed if and only if one can gather all the shared parts. It prevents semi-honest adversaries from other data holders. 
	\item[$\bullet$] TLS/SSL protocol ensures security and data integrity of pair-wise network communication \cite{aono2017privacy}. 
\end{itemize}	 


\subsubsection{Extra communication overhead}
N-out-of-n secret sharing leads to a quadratic growth of communication overhead with respect to the number of data holders. The overhead of aggregating gradient of local model for update is given as $\mathcal{O}(dWP^2)$, where $d$ denotes the data size of each weight.
In addition, as we have explained in the forward process, the local embedding and the global embedding are transmitted between server and data holder at each layer. Let $b$ denotes the length of node embedding, $N$ denotes the number of nodes from all local graphs, the communication overhead can be represented as $\mathcal{O}(bdNL)$. Therefore, although a small number of layers is sufficient for training a competitive GNN \cite{chen2020simple} that impedes over-smoothing, the communication overhead will become a bottleneck and limits efficiency and scalability, since $W$ and $N$ can be extremely large in the case of a heavy model with millions of parameters, or the  Internet of Things scenario with massive devices \cite{gao2020end}. Potential solutions include conducting mini-batch training instead of full-batch training, or utilizing model and communication compression technology \cite{rothchild2020fetchsgd}. We leave these optimizations as future works.

\subsubsection{Secure global pooling aggregation}
Note that when conducting global aggregation, only the element-wise maximum values over all local embedding in (\ref{rfg}) are required during forward step, while corresponding indexes of data holder (i.e., $\mathrm{P3}$ in (\ref{derivative0})) are needed at backward step. 
To further improve privacy, the raw information can be encrypted by private compare approaches by exploiting the technique of secure maximum computation protocol, which has been widely utilized in machine learning applications such as k-means \cite{jaschke2018unsupervised,mohassel2020practical}. Specifically, for each element of local embedding $t^{(p)}$, the problem of outputting the secret share of the index vector $\mathbf{I}=(0,\cdots,1,\cdots,0)$ that indicates the maximum value among $P$ numbers can be formulated as $\langle\mathbf{I}\rangle_{\oplus}=f_{max}(\{\langle t^{(p)}\rangle \mid {p\in\mathcal{P}}\})$. This function has been deeply investigated in recent works such as \cite{mohassel2020practical}, which can be efficiently implemented  by employing $P-1$ less-than garbled circuits and $4(P-1)$ instances of oblivious transfer extension.
Utilizing the secure global pooling aggregation
leads to more obstacles for the semi-honest server to learn private information from data holders.

\section{Evaluation}
In this section we present our experimental results for our proposed SAPGNN. We first describe the datasets, experimental setup and comparison methods. After that, we ran experiments to point the superiority of SAPGNN under a near IID label distribution setting. 
\begin{table}
	\caption{Main characteristics of each dataset}
	\centering
	\label{tab:data}
	\begin{tabular}{lrrr}
		\toprule
		Subgraph  &  Cora&  Citeseer & Pubmed    \\
		\midrule
		Nodes       & 2708  & 3327  & 19717  \\
		Edges       & 5278    &4552&44324\\
		Features    & 1433 &3703 &  500 \\
		Train & 140  &120 & 60  \\
		Val & 500 &500 & 500  \\
		Test & 1000  &1000 & 1000  \\
		Classes       & 7 &6 & 3\\
		\bottomrule
	\end{tabular}
\end{table}	
\subsection{Datasets and experimental setup}
We test SAPGNN on three publicly available citation node classification datasets that are used for node classification in previous works \cite{zhou2020privacy,zheng2021asfgnn}, i.e., Cora, Citeseer and Pubmed. For these datasets, each node represents a document, while edges denote citation links. Each node has a bag-of-words feature vector and a label indicating its category. We follow the same node mask with the default setting of DGL framework \cite{wang2019dgl} for training, validation, and test node sets. The main characteristics of each dataset are given in Table \ref{tab:data}. All experiments are evaluated on a Windows desktop with 3.2G 6-core Intel Core i7-8700 CPU and 16 GB of RAM. 


\subsection{ Compared methods}	 
We compare SAPGNN against two methods
\begin{itemize}
	\item[$\bullet$] The first is  separate training (\textbf{SP}), i.e., each data holder trains GNN individually over their own subgraph. It cannot utilize information from others and thus can be treated as a baseline method.
	\item[$\bullet$] The second is PPGNN \cite{zhou2020privacy} that first conducts separate training and then predicts over combined node embedding. Note that training, validation, and test node sets for PPGNN need to be privately aligned among data holders respectively before experiments since it requires each node exists at all local graphs.
\end{itemize}

\begin{figure*}
	\centering
	\includegraphics[width=0.99\linewidth]{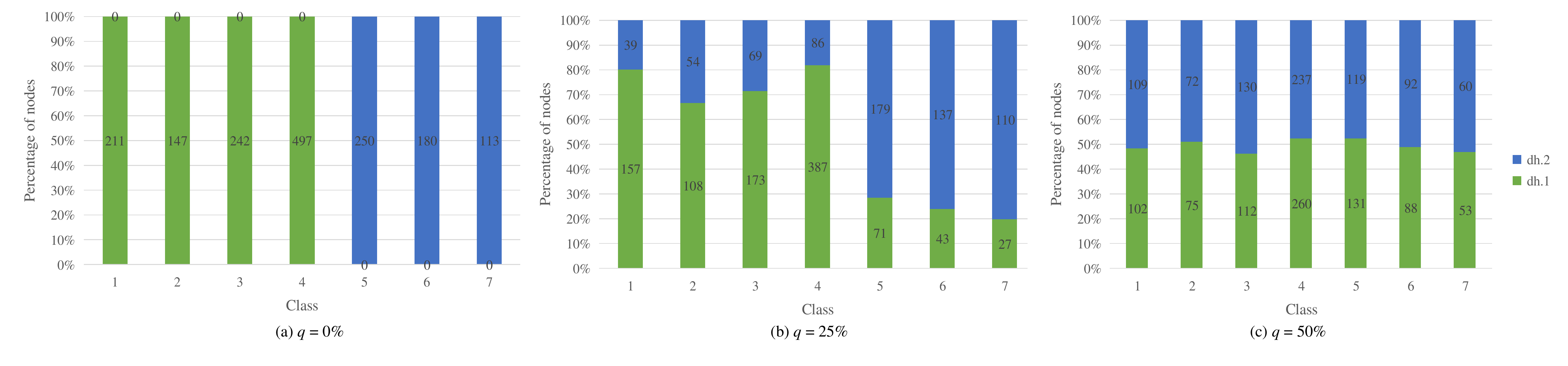}
	\caption{The percentage of nodes for each class with non-IID Cora dataset and two data holders when (a) $q$=0\%, (b) $q$=25\% and (c) $q$=50\%, where $q=0\%$ means each subgraph includes the nodes with different classes, while $q=50\%$ means each subgraph contains about half of nodes of each class.}
	\label{fig:fig3}
\end{figure*}

For all methods, we use a two-layer GNN  constructed by following formulation:
\begin{align}\label{a}
	\mathbf{h}_{v}^{(l+1)} = \mathbf{W}(\mathbf{h}_{v}^{(l)}+\mathrm{max}(\left\{\mathbf{h}_{u}^{(l)} \mid u \in \mathcal{N}(v)\right\})),
\end{align}
ReLU activation function and dropout are applied on the output of each layer except the last one. 
All the considered models are trained over a maximum of 300 epochs using the cross-entropy with Adam optimizer and learning rate of 0.01.
We performed a grid search with early stop to find the best choices for the hidden size for each method, and the accuracy and macro-F1 are evaluated on the test set over 40 consecutive runs.


\begin{table*}
	\caption{Comparison of accuracy and macro-F1 ($\pm$ standard deviation) over varying number of data holders from 1 to 4. The edges are divided uniformly into all data holders.
	}
\centering
	\label{tab:metric}
	\begin{tabular}{llcccccccc}
		\toprule
		&Number of data holders &1& & 2&& 3 &  &   4&\\
		Dataset &Model  & Acc & F1 & Acc & F1 & Acc & F1 & Acc & F1 \\
		\midrule
		Cora&\textbf{SP} &\textbf{78.5} &\textbf{77.4}& 75.2&74.3& 72.7& 71.7 & 70.6 &69.5 \\
		&&\textbf{$\pm$0.54}&\textbf{$\pm$0.55}&$\pm$0.79&$\pm$0.77&$\pm$0.85&$\pm$1.00&$\pm$0.87&$\pm$0.91\\
		&\textbf{PPGNN} &-- &--& 77.5&76.5&77.0&75.9&76.4& 75.1\\
		&&--&--&$\pm$1.36&$\pm$1.31&$\pm$1.10&$\pm$1.14&$\pm$1.36&$\pm$1.36\\
		&\textbf{SAPGNN}&\textbf{78.5} &\textbf{77.4}&\textbf{78.5} &\textbf{77.4}&\textbf{78.5} &\textbf{77.4} &\textbf{78.5} &\textbf{77.4} \\
		&&\textbf{$\pm$0.54}&\textbf{$\pm$0.55}&\textbf{$\pm$0.54}&\textbf{$\pm$0.55}&\textbf{$\pm$0.54}&\textbf{$\pm$0.55}&\textbf{$\pm$0.54}&\textbf{$\pm$0.55}\\
		Citeseer&\textbf{SP}&\textbf{69.8}&\textbf{66.6}&68.0&64.8&65.2 & 61.7  & 63.2 & 59.0 \\
		&&\textbf{$\pm$0.59}&\textbf{$\pm$0.62}&$\pm$1.81&$\pm$1.76&$\pm$0.99&$\pm$1.16&$\pm$2.12&$\pm$2.38\\
		&\textbf{PPGNN} &-- &--& 67.1&63.3&66.3&62.8&64.9&61.5\\
		&&--&--&$\pm$1.72&$\pm$2.45&$\pm$2.07&$\pm$1.81&$\pm$2.69&$\pm$2.58\\
		&\textbf{SAPGNN} &\textbf{69.8}&\textbf{66.6}&\textbf{69.8}&\textbf{66.6}&\textbf{69.8}&\textbf{66.6}&\textbf{69.8}&\textbf{66.6}\\
		&&\textbf{$\pm$0.59}&\textbf{$\pm$0.62}&\textbf{$\pm$0.59}&\textbf{$\pm$0.62}&\textbf{$\pm$0.59}&\textbf{$\pm$0.62}&\textbf{$\pm$0.59}&\textbf{$\pm$0.62}\\
		Pubmed&\textbf{SP}&\textbf{78.3}&\textbf{77.7}& 75.9&75.3&73.9&73.4&72.2&71.7\\
		&&\textbf{$\pm$0.51}&\textbf{$\pm$0.49}&$\pm$1.08&$\pm$1.12&$\pm$1.09&$\pm$1.08&$\pm$1.01&$\pm$1.01\\
		&\textbf{PPGNN} &-- &--&\textbf{78.9}&\textbf{78.4}&\textbf{79.0}&\textbf{78.7}&\textbf{79.2}&\textbf{79.0} \\
		&&--&--&\textbf{$\pm$0.88}&\textbf{$\pm$0.84}&\textbf{$\pm$0.58}&\textbf{$\pm$0.57}&\textbf{$\pm$0.61}&\textbf{$\pm$0.56}\\
		&\textbf{SAPGNN}&\textbf{78.3}&\textbf{77.7}&78.3&77.7&78.3&77.7&78.3&77.7\\
		&&\textbf{$\pm$0.51}&\textbf{$\pm$0.49}&$\pm$0.51&$\pm$0.49&$\pm$0.51&$\pm$0.49&$\pm$0.51&$\pm$0.49\\
		\bottomrule
	\end{tabular}
\end{table*}

\subsection{Results with uniformly split edges}
Firstly, we compare the three decentralized learning methods under the IID edge information setting, where the original edge set is divided uniformly into the subgraph of each data holder,
and the performance results are reported on Table \ref{tab:metric}. First, we can observe that the metrics of SAPGNN keep identical with varying numbers of data holders, and equal to the results obtained by centralized counterpart (i.e., SP when the number of data holders is 1). The reason is straightforward, as the learned global node representation of SAPGNN is the same as that learned over the combined graph. Secondly, SAPGNN consistently outperforms SP, and the gaps widen with the growth of data holders, since SP only accesses local information. Compared to PPGNN, SAPGNN is competitive for Cora and Citeseer datasets, but is slightly worse in the case of Pubmed.
In the following, we mainly compare SAPGNN and PPGNN in case of non-IID label distribution and drop the SP method for conciseness.

\subsection{Results with varies label distribution}

Existing works have demonstrated that the performance of decentralized learning method decreases with the raise of non-IID label distribution \cite{gao2020end,zhao2018federated}.
To examine this, we first divide nodes into different data holders according to the label, and then $q$\% nodes from each data holder are split uniformly to other data holders. Only the edges connected to nodes at the same data holder retained. Thus varying  the label distribution level $q$ from $0\%$ to $50\%$ implies more similar label distribution among data holders, and increasing the number of data holders will lead to more removed edges. Taking two data holders with Cora dataset as an example, Fig.\ref{fig:fig3} shows the percentage of nodes at different data holders for each class, where $q=0\%$ implies the labels among data holders are absolutely different, i.e., the subgraph at data holder 1 includes 1097 nodes of the first four classes, while data holder 2 only has 543 nodes with labels of the last three classes. As for the case of $q=50\%$, each subgraph contains about half of the nodes of each class (821 nodes at data holder1 while 819 nodes at data holder 2). Note that original PPGNN can only generate embeddings for overlapped nodes at all data holders. For fair comparison, instead of directly removing nodes, we remove all connected edges for these nodes at each local subgraph and thus no messages will pass from or to adjacent neighbors.
\begin{figure*}
	\centering
	\includegraphics[width=0.99\linewidth]{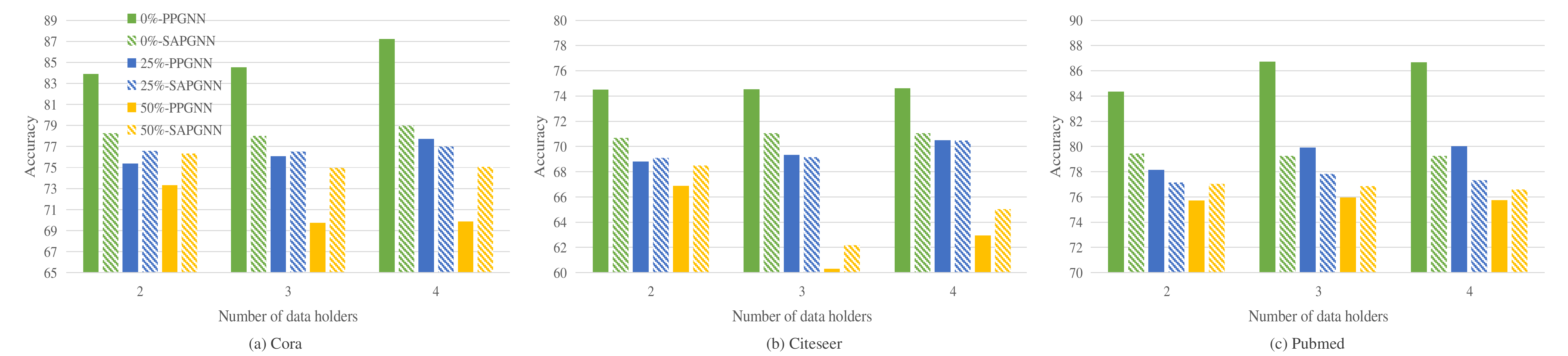}
	\caption{Node classification accuracy of SAPGNN and PPGNN, where the number of data holders is from 2 to 4 and $q\in\{0\%,25\%,50\%\}$.}
	\label{fig:fig4}
\end{figure*}
\begin{figure*}
	\centering
	\includegraphics[width=0.99\linewidth]{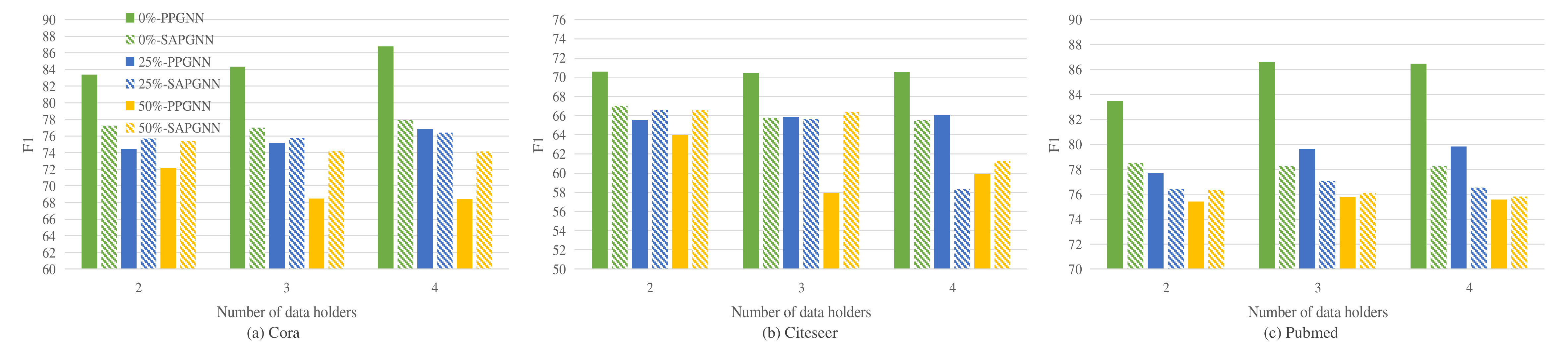}
	\caption{Node classification F1 of SAPGNN and PPGNN, where the number of data holders is from 2 to 4 and $q\in\{0\%,25\%,50\%\}$.}
	\label{fig:fig5}
\end{figure*}
Fig.\ref{fig:fig4} and Fig.\ref{fig:fig5} respectively show the node classification accuracy and F1 score when the number of data holders varies from 2 to 4 and $q\in\{0\%,25\%,50\%\}$. We can observe that label distribution has an important influence on metrics. In specific, when $q=0\%$, the performance of PPGNN  has a comfortable lead over SAPGNN.This is because PPGNN generates node embedding locally and thus can balance the contributions from different data holders. When the classes of nodes are totally different among data holders, training a shared or federal model has no benefit over that learns individually with a relatively simple classification task \cite{zhao2018federated}. On the other hand, SAPGNN has comparable performance when $q=25\%$, and outperforms PPGNN when $q=50\%$ in all datasets, which means SAPGNN is more effective on learning from adjacent information for the scenario where all data holders tend to have near IID label distribution ($25\%\leq q\leq50\%$).  At last, by comparing the performances of SAPGNN with the same $q$ over various number of data holders, we can find that removing inter-class edges may reduce the learning performance for Citeseer, while has relatively low influence on those for Cora and Pubmed.

\section{Conclusion}
In this paper, we proposed a server aided privacy-preserving GNN framework for the horizontally partitioned graph structure dataset. It enables the ability to generate the same node embeddings as the centralized GNN without revealing raw data. Therefore, proven concepts from the centralized one (e.g., convergence and generalization) can also be transferred to the proposed SAPGNN. For privacy concerns, we further give a secure global pooling aggregation mechanism that is capable of hiding raw local embeddings from  semi-honest adversaries.
We showed successful cases of SAPGNN on the node classification task especially when the labels of isolated datasets tend to have identical distribution, but it behaves worse than existing methods under highly skewed non-IID label distribution. This observation can be utilized for the guidance of choosing suitable decentralized learning paradigms according to the distribution of graph data.

In future, we would like to transfer our proposed learning framework to more general GNN architecture and more partition types of graph dataset. More importantly, how to enhance communication efficiency should pay attention to unleash the full potential of SAPGNN and other decentralized GNN learning approaches for the applications in reality.


%


\ifCLASSOPTIONcompsoc


\ifCLASSOPTIONcaptionsoff
  \newpage
\fi



\bibliographystyle{IEEEtran}
\bibliography{ijcai21_}
\end{document}